\documentclass[sigconf,authorversion,noacm]{acmart}
\usepackage[utf8]{inputenc}
\AtBeginDocument{%
	\providecommand\BibTeX{{%
			\normalfont B\kern-0.5em{\scshape i\kern-0.25em b}\kern-0.8em\TeX}}}

\usepackage{amsmath}
\usepackage{color}
\usepackage{enumitem}
\usepackage{tabularx}

\begin{document}
	
	\title{Generating artificial texts as substitution or complement of training data.}
	
	\author{Vincent Claveau}
	\email{vincent.claveau@irisa.fr}
	\orcid{0000-0002-3459-0550}
	\affiliation{%
		\institution{IRISA-CNRS}
		\streetaddress{Campus de Beaulieu}
		\city{Rennes}
		\country{France}
		\postcode{F-35042}
	}
	
	\author{Antoine Chaffin}
\email{antoine.chaffin@imatag.com}
\affiliation{%
	\institution{IRISA-CNRS and IMATAG}
	\streetaddress{Campus de Beaulieu}
	\city{Rennes}
	\country{France}
	\postcode{F-35042}
}

	\author{Ewa Kijak}
\email{ewa.kijak@irisa.fr}
\affiliation{%
	\institution{IRISA-Univ Rennes}
	\streetaddress{Campus de Beaulieu}
	\city{Rennes}
	\country{France}
	\postcode{F-35042}
}

	
	\begin{abstract}
		The quality of artificially generated texts has considerably improved  with the advent of transformers. The question of using these models to generate learning data for supervised learning tasks naturally arises. In this article, this question is explored under 3 aspects: (i) are artificial data an efficient complement? (ii) can they replace the original data when those are not available or cannot be distributed for confidentiality reasons? (iii) can they improve the explainability of classifiers? Different experiments are carried out on Web-related classification tasks  --- namely sentiment analysis on product reviews and Fake News detection ---  using artificially generated data by fine-tuned GPT-2 models. The results show that such artificial data can be used in a certain extend but require pre-processing to significantly improve performance. We show that bag-of-word approaches benefit the most from such data augmentation.
	\end{abstract}
	
	
	
	
	
	\maketitle

\section{Introduction}

Even if text generation is not a new technology, recent neural approaches based on transformers offers good enough performance to be used in various contexts \cite{Vaswani_NIPS2017}. 
In this paper, we explore the use of artificially generated texts for supervised machine learning tasks within two different scenarios: the artificial data is used as a complement of the original training dataset (for instance, to yield better performance) or the data is used as a substitute of the original data (for instance, when the original data cannot be shared because they contain confidential information \cite{Amin-NejadLREC2020}).
The generation of these artificial texts is performed with a neural language model trained on the original training texts.
In this paper, we show the interest of these scenarios with two Web-related text classification tasks, handling real and noisy language: fake news detection and opinion mining. 

Precisely, the main research questions dealt with in this paper are the following ones:
\begin{enumerate}
	\item what is the interest of text generation to improve text classification (complement);
	\item what is the interest of text generation to replace the original training data (substitution);
	\item what is the interest of text generation for explainable classifiers, based on bag-of-words representation.
\end{enumerate}

In the remaining of the paper, after a presentation of related work in Section~\ref{sec:connex}, we detail our classification approaches based on artificial text generation (Sec.~\ref{sec:approche}). The tasks and experimental data are described in Section~\ref{sec:data}. The experiments and their results for each of our research questions are reported in Section~\ref{sec:expes1} for the neural classifiers and Section~\ref{sec:expes2} for bag-of-words based classifiers.

\section{Related work}
\label{sec:connex}

Data augmentation for tasks of Natural Language Processing (NLP) has already been explored in several studies.
Some researchers propose more or less complex automatic modifications of the original examples in order to create new examples that are worded differently but are similar with respect to the NLP task (same class, same relation between words...). This is done for instance by simply replacing some words by  synonyms \cite{kobayashi-2018-contextual,wei-zou-2019-eda,Mueller2016,Jungiewicz2019}. The synonyms can be found in external resources such as WordNet \cite{wordnet95}, or in distributional thesauri, or from static word embeddings (such as Glove \cite{pennington2014glove} or word2vec \cite{Mikolov2013}).

In a similar vein since it only modifies the original examples locally, some neural techniques exploit masked language models (such as \textsc{Bert} \cite{Devlin2019}), that is, context-sensitive word embedding. 
These approaches works by masking a word in an original examples with the  \textsc{[mask]} token and to condition its replacement by a word from the expected class \cite{Wu2019}. 
It allows to generate a new example by replacing a word with another semantically close word (ideally a synonym). It is worth noting that, contrary to what we propose, the new example is not totally different (the syntactic structure of the new example is for instance very  similar to the original one).

Other approaches make the most of language models such as GPT-2 (Generative Pre-Trained Transformers \cite{Radford2019})  in order to produce a large quantity of data (texts) that are similar to the original data distribution. In Information Retrieval, this principle has been exploited to expand users' queries \cite{ClaveauRI2020}. 
Even closer, text generation is used for relation extraction \cite{Papanikolaou2020}, sentiment analysis of critics and questions \cite{Kumar2020} or for the prediction of hospital readmission and phenotype classification \cite{Amin-NejadLREC2020}. 
This paper is part of this line of work. 
Our interest here is to examine the gains and losses of our different scenarios of using artificial data, their preparation, and to examine their effects on different families of classifiers.

\section{Generating artificial data}
\label{sec:approche}

\begin{figure*}
	\centering
	\includegraphics[width=0.7\linewidth]{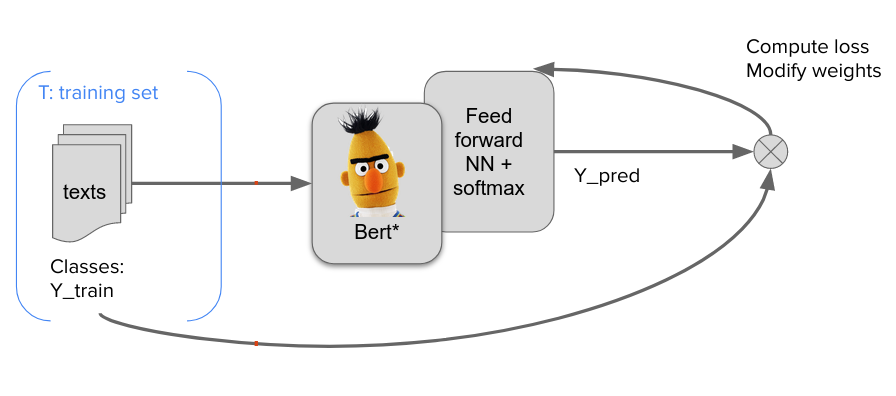}
	\includegraphics[width=0.8\linewidth]{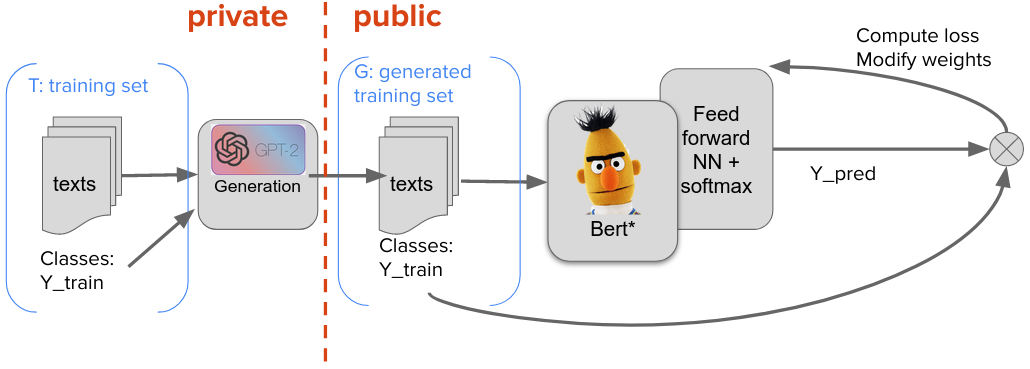}
	\includegraphics[width=0.8\linewidth]{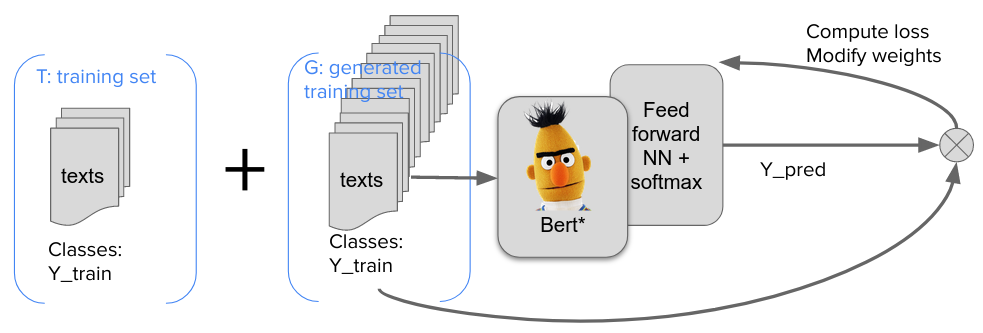}
	\caption{Different training scenarios: usual text classification framework (here with a standard \textsc{Bert} classifier), generated data used as substitution (especially useful when the original data cannot be shared), generated data as complement.}
	\label{fig:scenarios}
\end{figure*}

Let us assume to have a set of (original) texts ${\mathcal T}$ divided into $n$ classes $c_i$, from which we wish to generate artificial texts ${\mathcal G}_{c_i}$ for each class $c_i$.
As explained in the introduction, we want to examine different scenarios of usage of these generated data: complement or substitution. The scenario, as well as the usual text classification framework, are examplified in Figure~\ref{fig:scenarios}.

We use GPT models to generate the artificial texts. These models are built by stacking \textit{transformers} (more precisely decoders), trained on large corpora by auto-regression, i.e. on a task of predicting the next word (or \textit{token}) knowing the previous ones.
The second version, GPT-2~\cite{Radford2019}, contains 1.5G parameters for its largest model, trained on more than 8 million documents from Reddit (i.e. general domain language such as discussions on news articles, mostly in English).
		
A newer version, GPT-3, has been released in July 2020; it is much more larger (175 billion parameters) and outperforms GPT-2 on any tested task. Yet, the experiments reported below needs fine-tuning, which is not feasible with such a large model which rather rely on prompt engineering for task adaptation.

\subsection{Fine-tuning the language model.}

For this fine-tuning step, we start from the medium model (774M parameters) pre-trained for English and made available by OpenAI\footnote{\url{https://github.com/openai/gpt-2}}. 

In the work presented in this paper, we fine-tune one language model per class with the original training data ${\mathcal T}$.
Another training procedure available in the literature is to adapt a single model, but to condition it with a special \textit{token} indicating the expected class at the beginning of the text sequence (i.e. at the beginning of each original example). 
Due to the limited amount of data available per class (compared to the number of parameters of the GPT-2 model), it is important to control the fine-tuning to avoid overfitting. To do so, we limit the number of epochs to 2,000; the other fine-tuning parameters are the default ones of the OpenAI GPT2 code that is used in our experiments. On a Tesla V100 GPU card, this fine-tuning step lasts about 1 hour for each dataset (see below).

\subsection{Text generation.}

For each class ${c_i}$ of the dataset ${\mathcal T}$, we use the corresponding model to generate artificial texts ${\mathcal G}_{c_i}$ which hopefully will fall into the desired class. 
We provide prompts for these texts in the form of a start-of-text token followed by a word randomly drawn from the set of original texts.
Several parameters can influence the generation. We used the default values that we give here for reproducibility purposes, without detailing them (see the GPT-2 documentation): 
\texttt{temp.} = 0.7, \texttt{top\_p} = 0.9, \texttt{top\_k} = 40.

The texts generated for the class $c_i$ containing a sequence of 5 consecutive words appearing identically in a text of ${\mathcal T}_{c_i}$ are removed. This serves two purposes~: on the one hand, it limits the risk of revealing an original document in the case where the ${\mathcal T}_{c_i}$ data are confidential, and on the other hand, it limits the duplicates which are harmful to the training of a classifier in the case where the ${\mathcal G}_{c_i}$ data are used in addition to ${\mathcal T}_{c_i}$.
In practice, this concerns about 10\% of the generated texts in our experiments. 
Note that in the scenario where the data are confidential, providing the generator itself is not possible, since it may be used to find back the data it was fine-tuned on.
In the experiments reported below, 16,000 texts are thus generated for each $c_i$ class (this number of texts has been fixed arbitrarily).

\subsection{About confidentiality}

In the scenario where the original data cannot be distributed, notably for confidentiality reasons, it is appropriate to ask whether sensitive information can be recovered with the proposed approach.
As said earlier, if the whole generative model is made available, this risk has been studied \cite{Carlini2020}, and exists, at least from a theoretical point of view under particular conditions\footnote{See also the discussion on the Google AI blog: \url{https://ai.googleblog.com/2020/12/privacy-considerations-in-large.html}.}.

When only the generated data are made available, there is also a risk of finding confidential information in them. 
Without other safeguards, it is indeed possible that among the generated texts, some are paraphrases of sentences of the training corpus. 
However, in practice, the risk is very limited:
\begin{itemize}
	\item first of all, because there is no way for the user to distinguish these paraphrases among all the generated sentences;
	\item secondly, because additional measures can be taken upstream (for example, de-identification of the training corpus) and downstream (deletion of generated sentences containing specific or nominative information...); 
	\item Finally, more complex systems to remove paraphrases, such as those developed for the  \textit{Semantic Textual Similarity} tasks \cite[\textit{inter alia}]{Jiang2020}, can even be considered. 
\end{itemize}
These measures make it highly unlikely that any truly usable information can be extracted from the generated data.

\section{Classification tasks and datasets}
\label{sec:data}

The experiments detailed in the next section are real classification tasks of the Web: fake news detection in tweets and sentiment analysis in reviews. They are both classification tasks usually dealt with by machine learning. We test different languages: the fake news dataset consists of tweets in English while the sentiment analysis dataset is in French. They are presented hereafter. 

\subsection{FakeNews MediaEval 2020: English dataset of tweets}

This dataset was developed for the detection of fake news within social networks as part of the MediaEval 2020 FakeNews challenge \cite{pogorelov2020fakenews}.
In this task, tweets about 5G or coronavirus were manually annotated according to three classes $c_i, i \in \{'5G', 'other', 'non'\}$ \cite{schroeder2019fact}. $'5G'$ contains tweets propagating conspiracy theories associating 5G and coronavirus, $'other'$ are for tweets propagating other conspiracy theories (which may be about 5G or covid but not associated), and $'non'$ tweets not propagating any conspiracy theories.

It is worth noting that the classes are imbalanced; indeed, in the training dataset ${\mathcal T}$~: $|{\mathcal T}_{5G}| = 1,076, |{\mathcal T}_{\textrm{other}}| = 620, |{\mathcal T}_{\textrm{non}}| = 4,173$.

The data augmentation (i.e., text generation) is performed as explained in the previous section.
Figure~\ref{fig:example_5G} presents three examples of generated texts from the MediaEval 2020 training for the '5G' class. 

\begin{figure*}
    \centering
    \fbox{%
    \parbox{1.4\columnwidth}{%
        \texttt{ 
        \begin{itemize}[leftmargin=*]
\item[-] If the FBI ever has evidence that a virus or some other problem caused or contributed to the unprecedented 5G roll
 out in major metro areas, they need to release it to the public so we can see how much of a charade it is when you try  to downplay the link.\\
\item[-] So let's think about this from the Start. Is it really true that 5G has been activated in Wuhan during Ramadan? Is this  a cover up for the fact that this is the actual trigger for the coronavirus virus? Was there a link between 5G and the coronavirus in the first place? Hard to say.\\
\item[-] We don't know if it's the 5G or the O2 masks that are killing people. It's the COVID19 5G towers that are killing people. And it's the Chinese people that are being controlled by the NWO
\end{itemize}}}}
    \caption{Examples of tweets artificially generated with a GPT-2 model trained on the MediaEval examples with class ${\mathcal T}_{5G}$.}
    \label{fig:example_5G}
\end{figure*}

\subsection{FLUE CLS-FR: French dataset for sentiment analysis}

The second dataset is taken from the FLUE evaluation suite for French \cite{Flaubert2020}.
It is the French part of the Cross Lingual Sentiment (CLS-FR) dataset \cite{Prettenhofer2010}, which consists of product reviews (books, DVD, music) from Amazon. The task is to predict whether the review is positive (rated more than 3 stars on the merchant site) or negative (less than 3 stars). The dataset is divided into balanced training and test sets. 
In our experiments, we do not distinguish between products~: we have only two classes (positive, negative) with  reviews of books, DVDs or music. 

As with the MediaEval data, a language model is tuned for each class using the training data. 
Generation is then done as described in the previous section. Examples of generated negative reviews are given in Figure~\ref{fig:example_CLS_neg}.

\begin{figure*}
	\centering
	\fbox{%
		\parbox{1.4\columnwidth}{%
			
				\begin{itemize}[leftmargin=*]
					\item[-] \texttt{  Déçue... J'ai eu je l'avoue du mal à lire ce livre arrivé au milieu de celui-ci. L'histoire ne paraît pas vraiment très réaliste. Le policier est plus guidé par de la chance que par son instinct. Que se serait-il passer s'il n'avait pas rencontré cette dame insolite ? Non ! Je ne crois pas que je lirais d'autres livres de fred vargas... Dommage je n'ai pas encore trouvé une source infaillible de bons polars.}\\
					\textit{translation: Disappointed... I have to admit that I had a hard time reading this book until the middle of it. The story doesn't really seem very realistic. The policeman is guided more by luck than by his instinct. What would have happened if he had not met this unusual lady? No! I don't think I would read any more books by fred vargas... Too bad I haven't found an infallible source of good thrillers yet.\\}
					\item[-]  \texttt{ De la daube. Cet homme ferait mieux de mettre son piano à la benne. Il n'y a pas de musicalité, ce disque irrite et agresse, ou au mieux il agresse et abuse son timbre et pénible accent amoureux. Musicalement, c'est de la musique de... chandler, on se dit... "c'mere irons up". Une chose est surement restée disponible sur cet album, mais attention aux maisons de disque !} \\
					\textit{translation: Rubbish. This man would do better to put his piano in the garbage. There is no musicality, this record irritates and assaults, or at best he assaults and abuses his timbre and painful love accent. Musically, it is music of... chandler, we say to ourselves... "c'mere irons up". A thing surely remained available on this album, but attention to the record companies!\\}
					\item[-]  \texttt{ Gros navet. Décor atrocement kitch, couleurs d'un mauvais goût abominable qui rendrait effleuré un ami en le dire... ça marche. Aucun suspense, tout est répétitif, les personnages sont inconséquents, ennuyeux. A eviter absolument.\\}
				\textit{translation: Such a turkey. Atrociously kitsch decor, colors of an abominable bad taste that would make a friend shudder to say it. No suspense, everything is repetitive, the characters are inconsistent, boring. To avoid at all costs.	}
	\end{itemize}}}
	\caption{Examples of artificial reviews generated with a GPT-2 trained on the CLS-FR examples with the class ${\mathcal T}_{negatif}$.}
	\label{fig:example_CLS_neg}
\end{figure*}

As can be seen from these examples (including the MediaEval examples in Figure~\ref{fig:example_5G}), the generated texts seem to belong to the expected class (see Section~\ref{sec:res2} for a discussion of this point). However, they often have flaws that make the fact that they were generated detectable. This is particularly the case for French texts, which can be explained by the fact that we did not have, at the time of the experiments, a pre-trained model for French; the model, as well as the tokenizer, are therefore based on the English GPT model. 
GPT-2 models for French released very recently\footnote{For example, the Pagnol model of LightOn: \url{https://lair.lighton.ai/pagnol/}.} could improve this aspect.

\section{Experiments: neural classification approaches}
\label{sec:expes1}

In the experiments reported below, the performance is measured in terms of micro-F1 (equivalent to accuracy), and, to take into account the imbalance of the classes (notably in the MediaEval dataset), in terms of macro-F1 and MCC (Matthews Correlation Coefficient\footnote{Also called $\Phi$~ coefficient; see \href{https://en.wikipedia.org/wiki/Matthews_correlation_coefficient}{the dedicated Wikipedia page}.}), as implemented in the library \href{https://scikit-learn.org}{scikit-learn}~\cite{scikit-learn}. The performance is measured on the official test sets of the MediaEval \cite{pogorelov2020fakenews} and CLS-FR \cite{Flaubert2020} tasks, of course disjoint from the training sets ${\mathcal T}$.

\subsection{First results}
\label{sec:res1}

For our first experiments, we use state-of-the-art neural classification models based on transformers. 
For the MediaEval data, in English, we opt for a Ro\textsc{bert}a \cite{roberta} pre-trained model for English (\textit{large} model with a classification layer). It is this type of transformer-based models that obtained the best results on these data during the MediaEval 2020 challenge \cite{CheemaMediaEval2020,ClaveauMediaEval2020}. Among the variants of \textsc{bert} \cite{Devlin2019}, Ro\textsc{bert}a was preferred here for its tokenizer that is more adapted to the specifics of the very free form of writing found in tweets (mix of upper and lower case, absence or multiplication of punctuation, abbreviations...). 
For the CLS-FR data of FLUE, we use the large-cased Flau\textsc{bert} model \cite{Flaubert2020}. This allows us to compare with the results originally published on these data.

We evaluate the performance according to our different training scenarios: on the original data ${\mathcal T}$ (which serves as a \textit{baseline}), on the artificial data ${\mathcal G}$, and finally on both the artificial and original data. 
In this last case, we test two training strategies~:
\begin{itemize}
	\item the first, ${\mathcal T} + {\mathcal G}$, mixes the original and artificial examples,
	\item the second, ${\mathcal G} \mbox{ then } {\mathcal T}$, trains on the artificial data on the first epochs, then on the original data for the last epoch. This results in a kind of fine-tuning on the original data after a first training on the artificial data.
\end{itemize} 
The implementation that we use is based on the HuggingFace's Transformers library \cite{HuggingFace} with the batch size set to 16 and the number of \textit{epochs} set to 3 in all scenarios (optimal number of epochs for the baseline), except the last one (3 on ${\mathcal G}$ followed by 1 on ${\mathcal T}$).

\begin{table*}[tb]
    \centering
    \begin{tabular}{l|ccc|ccc}
    	      & \multicolumn{3}{c|}{MediaEval} & \multicolumn{3}{c}{CLS-FR} \\
    model    &  micro-F1 &  macro-F1 & MCC                     & micro-F1 &  macro-F1 & MCC\\
    \hline
    \textsc{bert}* / ${\mathcal T}$             & 79,57 & 62,66 & 55,71      & 95,44   &  95,42  & 90,86 \\
    \hline
    \textsc{bert}* / ${\mathcal G}$             & 62,68 & 54,03 & 39,27      & 95,13   &  95,12  & 90,25 \\
    \textsc{bert}* / ${\mathcal T} + {\mathcal G}$  & 75,01 & 58,81 & 46,37      & 95,43   &  95,42  & 90,89 \\
    \textsc{bert}* / ${\mathcal G}$ puis ${\mathcal T}$  & 79,89 & 60,64 & 52,02 & 95,76   &  95,75  &  91,51 \\
    \end{tabular}
    \caption{Performance (\%) of neural classification techniques on data from MediaEval and CLS-FR according to the scenario of usage of the generated texts (without filtering) (cf. Sec.~\ref{sec:res1}). The \textsc{bert}* model are respectively Ro\textsc{bert}A and Flau\textsc{bert}. }
    \label{tab:res1}
\end{table*}

The results for the MediaEval and CLS-FR datasets are reported in Table~\ref{tab:res1}. 
On the CLS-FR data, we observe very few differences between the different scenarios and compared to the baseline (note that our baseline is similar to the published state-of-the-art results). 
The classification task, which is relatively simple, obviously generates data of as good quality as the original data, leading to comparable results. On this type of task, artificially generated data can therefore be used without loss of performance.

The MediaEval task is more difficult as can be seen with the results of the baseline (Ro\textsc{bert}A / ${\mathcal T}$). 
On these data, in a substitution scenario (i.e. when the generated data are used alone as training data), the results are strongly degraded compared to a system trained on the original data. 
This is of course due to the fact that the data generated by each of the language models may not belong to the expected class, as the models do not fully capture the specificity of the \textit{fine-tuning} data.
In a learning data complement scenario, the impact is less significant, especially if the artificial data is used only on the first few epochs.

\subsection{Results with automatic filtering}
\label{sec:res2}

As we have seen, the ${\mathcal G}$ examples generated by our trained GPT-2 models may contain texts that do not belong to the expected classes. Manually filtering or annotating these texts is of course possible but remains a costly task. 
To reduce the effect of these texts on the classification at a lower cost, we propose to exclude them using a first classifier learned on the original data ${\mathcal T}$: any text of ${\mathcal G}_{c_i}$ which is not classified $c_i$ by the classifier is removed. In this way, we hope to eliminate, automatically, the most obvious cases of problematic artificial texts.
In the following experiments, we use the Ro\textsc{bert}a classifier trained on ${\mathcal T}$ (evaluated in the first row of Tab.~\ref{tab:res1}). 
In this way, 40\% of the examples are deleted. The artificial examples kept are noted ${\mathcal G}^f$.

The results with these new filtered sets of artificial examples in the same training scenarios are presented in Table~\ref{tab:res2} for the MediaEval and CLS-FR data.
It can be seen that this filtering strategy pays off, with improved performance on all metrics compared to no filtering. In the substitution scenario, the performance is now close to the \textit{baseline}, and is even better on the macro-F1; this is explained by the fact that the artificial set $\mathcal G$ is much more balanced than $\mathcal T$ and thus performs better on the minority classes of the test set. In the complement scenario, we observe a significant improvement over the baseline, especially with the sequential strategy.

\begin{table*}[tb]
	\centering
	\begin{tabular}{l|ccc|ccc}
		 & \multicolumn{3}{c|}{MediaEval} & \multicolumn{3}{c}{CLS-FR} \\
		model    &  micro-F1 &  macro-F1 & MCC                        &  micro-F1 &  macro-F1 & MCC \\
		\hline
		\textsc{bert}* / ${\mathcal T}$                    & 79,57 & 62,66 & 55,71  & 95,44  & 95,42  & 90,86 \\
		\hline
		\textsc{bert}* / ${\mathcal G}^f$                  & 76,22 & 64,18 & 52,75  & 95,76  & 95,75  & 91,51 \\
		\textsc{bert}* / ${\mathcal T} + {\mathcal G}^f$       & 80,12 & 66,08 & 57,44  & 95,99  & 95,98  & 91,97 \\
		\textsc{bert}* / ${\mathcal G}^f$ puis ${\mathcal T}$  & 83,55 & 67,90 & 60,05  & 95,96  & 95,95  & 91,96\\
	\end{tabular}
	\caption{Performance (\%) or neural classification approaches on the MediaEval et CLS-FR tasks according to our scenarios of usage of the artificially generated texts after filtering (cf. Sec.~\ref{sec:res2}). The \textsc{bert}* model are respectively Ro\textsc{bert}A and Flau\textsc{bert}..}
	\label{tab:res2}
\end{table*}

\subsection{Differences between classifiers}

Beyond the global performance measures, it can be interesting to check if the classifier trained on the artificial data allows to make the same decisions as a classifier trained on $\mathcal T$. To do so, we can look at the proportion of examples (from the test set) for which the decision between \textsc{bert}* / ${\mathcal T}$ and \textsc{bert}* / ${\mathcal G}^f$ differs.
For the CLS-FR data, the classifiers agree on a large majority of examples. Figure~\ref{fig:cm} shows the confusion matrix of Flau\textsc{bert} / ${\mathcal T}$ and Flau\textsc{bert} / ${\mathcal G}^f$ on the CLS-FR data.

\begin{table*}[h!]
	\begin{center}
		\begin{tabular}{ m{3cm} | m{5cm} | m{5cm}  }
			
			 & ~ ~ ~ ~ ~ ~ Positive (ground-truth) & ~ ~ ~ ~ ~ ~ Negative (ground-truth)\\ 
	\hline
Predicted positive &	~\newline \includegraphics[width=0.7\linewidth]{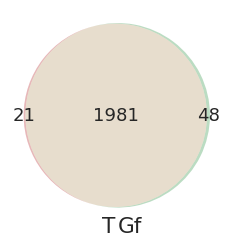}  & ~\newline  \includegraphics[width=0.7\linewidth]{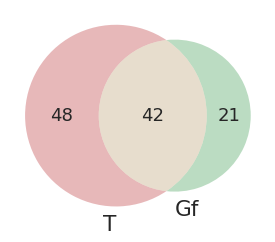} \\
\hline
Predicted negative	&  ~\newline  \includegraphics[width=0.7\linewidth]{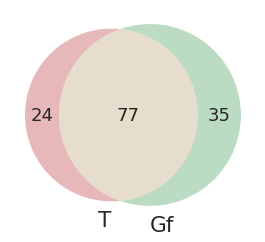}  & ~\newline  \includegraphics[width=0.7\linewidth]{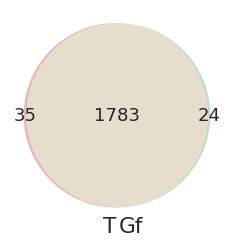} \\
\end{tabular}
	\caption{Confusion matrix of the Flau\textsc{bert} / ${\mathcal T}$ et Flau\textsc{bert} / ${\mathcal G}^f$ models on the CLS-FR data. The Venn diagrams shows the proportions of shared examples for each category.}
	\label{fig:cm}
\end{center}
\end{table*}

From this confusion matrix, we can see that the classifiers do agree on the majority of examples. 
The cases of disagreement are proportionally more important on the false positives and false negatives, but even for these categories, we still find a lot of common errors (42 and 77 examples respectively for the false positives and false negatives).
The classifiers have therefore not only comparable performance, but very similar behaviors in detail since they give the same class on most examples.

\section{Experiments: bag-of-words approaches}
\label{sec:expes2}

We also test classifiers based on bag-of-words representations; we present only the results of the logistic regression (LR) which gave the best results. In general, these classifiers perform less well than the transformers-based approaches, but they allow for better explainability \cite[for a definition and characterization of learning methods]{Miller2017,Carvalho2019}, for example by examining the regression weights associated with words. They are also way less expensive to train.

\subsection{First results}

The implementation used is scikit-learn~\cite{scikit-learn}, the texts are vectorized with TF-IDF weighting and L2-normalized, and the LR parameters are the default ones except for the following~: multiclass strategy \textit{one-vs.-rest}, number of iterations = 2500.
Results for the same scenarios as above are presented for the MediaEval and CLS-FR tasks in Tables~\ref{tab:res3-mediaeval} and \ref{tab:res3-CLS-FR}.

\begin{table}[tb]
    \centering
    \begin{tabular}{l|ccc}
    model    &  micro-F1 &  macro-F1 & MCC \\
    \hline
    LR  / ${\mathcal T}$                   & 72,68 & 56,35 & 42,22 \\
    LR / ${\mathcal G}^f$                  & 74,00 & 59,18 & 44,39 \\
    LR / ${\mathcal T} + {\mathcal G}^f$       & 75,46 & 59,64 & 45,83 \\
    \end{tabular}
    \caption{Performance (\%) of the LR/bag-of-words approach on the MediaEval dataset according to our scenarios of usage of the artificially generated data after filtering: without, substitution, complement.}
    \label{tab:res3-mediaeval}
\end{table}

\begin{table}[tb]
	\centering
	\begin{tabular}{l|ccc}
		model    &  micro-F1 &  macro-F1 & MCC \\
		\hline
		LR  / ${\mathcal T}$                   & 84,77 & 84,70 & 69,48 \\
		LR / ${\mathcal G}^f$                  & 87,16 & 87,14 & 74,27 \\
		LR / ${\mathcal T} + {\mathcal G}^f$       & 88,36 & 88,34 & 76,69 \\
	\end{tabular}
	\caption{Performance (\%) of the LR/bag-of-words approach on the CLS-FR dataset according to our scenarios of usage of the artificially generated data after filtering: without, substitution, complement.}
	\label{tab:res3-CLS-FR}
\end{table}

For this type of classifier, the interest of the generated data appears for both scenarios and on the two datasets. 
In the case of substitution, the classifiers are slightly better than those trained on the original data. 
This demonstrates the importance of having a larger amount of data to capture form variants in texts (synonyms, paraphrases...) that the bag-of-words representations cannot otherwise capture as easily as the pre-trained embedding-based representations of the \textsc{Bert} models.
In the scenario where data is used as a complement, the performance increase is even more marked and thus becomes close to the neural baseline, while having the advantages of a classifier considered more interpretable.

\subsection{Impact of the quality of the generated data}

\begin{figure*}
	\centering
	\includegraphics[width=0.65\linewidth]{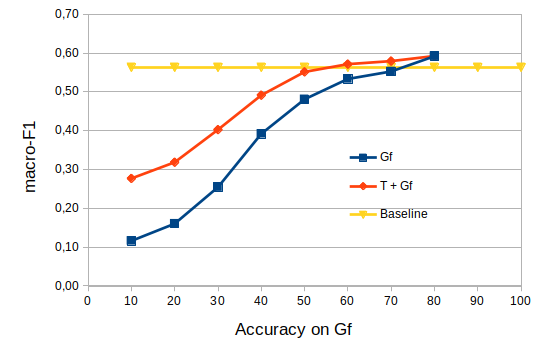}
	\caption{Performance (macro-F1) according to the quality (accuracy in \%) of the classifier filtering the artificially generated data; MediaEval dataset with logistic regression.}
	\label{fig:bruit}
\end{figure*}

It is interesting to examine what is the influence of the quality of the generated data (even filtered) on the results of the final classifier (see Section~\ref{sec:res2}).
To study this, we inject noise into the classification to simulate filtering done with classifiers of varying quality.
This is done simply by replacing, for randomly drawn examples of ${\mathcal G}^f$, the predicted class (by the generator and by the filtering classifier) by a randomly drawn class. The number of examples undergoing this treatment is computed so that the errors inserted make the accuracy of the dataset drops to 80,\%, 70,\%, etc. 
The effect of these errors in the generated examples on the final performance of the complement and substitution strategies are presented in Figure~\ref{fig:bruit} (MediaEval data) with logistic regression as final classifier.

As can be seen in this figure, empirical results about the influence of filtering quality are unsurprising. In the substitution scenario, the final performance is strongly dependent on the quality of the filtering classifier; in this case, a performance level equivalent to the original dataset is achieved when the accuracy of the filter exceeds 70\,\%. In the case of the complement scenario, the gain is significant as soon as the filter has an accuracy higher than random.


\section{Concluding remarks}

In this work, we have explored the interest of text generation for two text classification tasks from the Web (fake news detection in tweets and sentiment analysis on product reviews).
In a scenario where the original training data cannot be distributed, we have shown that it is possible to generate artificial data for supervised learning purposes. For state-of-the-art classifiers based on transformers, this degrades the performance (compared to the one achieved with the original data) but in a contained proportion (-4\% accuracy). On the other hand, for the classifiers exploiting bag-of-words representations, we notice an improvement due to the larger amount of training data available.

In a scenario where artificial data is added to the original data, we have shown that classifiers benefit from additional data, including neural networks. This result is particularly positive for the bag-of-words approaches, which are more sensitive to reformulations, and which clearly benefit from the addition of these artificial examples. We thus have a good compromise between methods that are fast to train, more easily explainable, while having performance close to neural networks. 

As we have seen, these results are obtained provided that the generated data are filtered beforehand, which seems to contradict several studies cited in Sec.~\ref{sec:connex}. In our experiments, this was done automatically; manual correction of the data (of their classes) is also possible and may allow better results, but with an additional annotation cost.
The use of these methods for other data and other NLP tasks than text classification remains a promising avenue. Among these NLP tasks, those based on word labeling (token classification) pose different problems and require adapted solutions.
In the future, it would be interesting to verify the consistency of our results according to other generation approaches \cite{Kumar2020}. 
It also seems interesting to study more deeply the impact of the quality of the classifier used to filter the artificial data. Moreover, the integration of this filtering step as a constraint during the generation of artificial examples is a promising avenue.

For replicability purposes, the training scenarios presented in this article are available online for the MediaEval\footnote{\url{https://colab.research.google.com/drive/1VDm-MZcgVJpMaVmmGa1Wvmyc71q6IYBJ}} and CLS-FR\footnote{\url{https://colab.research.google.com/drive/1i2IOBV5yEi2ID9atyMBn6PdX5xnschtK}} tasks. 
The generation of examples relies on \url{https://github.com/minimaxir/gpt-2-simple}. The data are available from their producers (see Section~\ref{sec:data}).



\bibliographystyle{ACM-Reference-Format}
\bibliography{biblio_gpt4classif}

\end{document}